\documentclass{article}

\usepackage{microtype}
\usepackage{graphicx}
\usepackage{subfig}  
\usepackage{booktabs} 

\graphicspath{{figures/}}  

\usepackage{hyperref}

\usepackage[accepted]{icml2021}

\icmltitlerunning{Inverse problems by DNNs driven by physics-based dictionary}

\usepackage{mathtools} 										
\usepackage[binary-units=true]{siunitx} 	
\usepackage{placeins} 						
\usepackage{amsfonts}  

\renewcommand{\v}[1]{\ensuremath{\mathbf{#1}}} 				
\newcommand{\gv}[1]{\ensuremath{\mbox{\boldmath$ #1 $}}} 	
\DeclareMathOperator*{\argmin}{argmin}						
\newcommand{\Ntot}[0]{\ensuremath{N_{\textrm{tot}}}}
\newcommand{\BigO}[1]{\ensuremath{\mathcal{O}\left( #1 \right)}}

\begin{document}
\twocolumn[
\icmltitle{Solving inverse problems with deep neural networks driven by sparse signal decomposition in a physics-based dictionary}


\icmlsetsymbol{equal}{*}  

\begin{icmlauthorlist}
\icmlauthor{Gaetan Rensonnet}{icteam}
\icmlauthor{Louise Adam}{icteam}
\icmlauthor{Benoit Macq}{icteam}
\end{icmlauthorlist}

\icmlaffiliation{icteam}{ICTEAM Institute, Universit\'{e} catholique de Louvain, Louvain-la-Neuve, Belgium}

\icmlcorrespondingauthor{Gaetan Rensonnet}{gaetan.rensonnet@uclouvain.be}

\icmlkeywords{Machine Learning, inverse models, interpretability, deep learning, fingerprinting, dictionary, non-negative linear least squares, multi-layer perceptron, brain, white matter, diffusion MRI, expert knowledge}

\vskip 0.3in
]



\printAffiliationsAndNotice{}  

\begin{abstract}

Deep neural networks (DNN) have an impressive ability to invert very complex models, i.e. to learn the generative parameters from a model's output. Once trained, the forward pass of a DNN is often much faster than traditional, optimization-based methods used to solve inverse problems. This is however done at the cost of lower interpretability, a fundamental limitation in most medical applications. We propose an approach for solving general inverse problems which combines the efficiency of DNN and the interpretability of traditional analytical methods. The measurements are first projected onto a dense dictionary of model-based responses. The resulting sparse representation is then fed to a DNN with an architecture driven by the problem's physics for fast parameter learning. Our method can handle generative forward models that are costly to evaluate and exhibits similar performance in accuracy and computation time as a fully-learned DNN, while maintaining high interpretability and being easier to train. Concrete results are shown on an example of model-based brain parameter estimation from magnetic resonance imaging (MRI).
\end{abstract}

\section{Introduction}
\label{sec:introduction}
Many engineering problems in medicine can be cast as inverse problems in which 
a vector of latent biophysical characteristics $\gv{\omega}$ (e.g., biomarkers) are estimated 
from a vector of $M$ noisy measurements or signals $\v{y}\in \mathbb{R}^M$, each measurement being performed with 
experimentally-controlled acquisition parameters $\v{p}_i$ ($i=1,\dots,M$) defining the experimental protocol $\mathcal{P} \coloneqq \left\{\v{p}_i\right\}_{i=1}^{M}$. 

We consider the case in which a generative model of the signal $S\left(\gv{\omega};\v{p}\right)$ is available but hard to invert and costly to evaluate. This occurs for instance when $S$ is very complex or non-differentiable, which includes most cases of $S$ being a numerical simulation rather than a closed-form model. Without loss of generality, the signal $S$ is assumed to arise from $K$ independent contributions weighted by weights $\nu_1,\dots,\nu_K$
\begin{equation}
S\left(\gv{\omega};\mathcal{P}\right) = \sum\limits_{k=1}^K \nu_k S_k\left(\gv{\omega}_k;\mathcal{P}\right),
\label{eq:linearity_S}
\end{equation}
where the $S_k$ are available and with $\gv{\omega}=\left[\gv{\omega}_1^T \dots  \gv{\omega}_K^T \right]^T$. When no linear separation is possible or known then $K=1$.


Because deep neural networks (DNNs) have shown a tremendous ability to learn complex input-output mappings \cite{hornik1990universal,fakoor2013using, guo2016deep,lucas2018using,zhao2019object}, it is tempting to resort to a DNN to learn $\gv{\omega}$ directly from the data $\v{y}$ or from synthetic samples $S\left(\gv{\omega};\mathcal{P}\right)$. This is arguably the least interpretable approach and is referred to as the fully-learned, black-box approach. Little insight into the prediction process is available, which may deter medical professionals from adopting it. In addition, changes to the input such as the size $M$ (e.g., missing or corrupt measurements) or a modified experimental protocol $\mathcal{P}$ (e.g., hardware update) likely require a whole new network to be trained.

At the other end of the spectrum, the traditional, physics-based approach in that case is to perform dictionary fingerprinting. As described in Section~\ref{sec:fingerprinting}, this roughly consists in pre-simulating many ($\Ntot{}$) possible signals $S$ and finding the combination of $K\ll \Ntot{}$ responses that best explains the measured data $\v{y}$. This method is explainable and interpretable but it is essentially a brute-force discrete search. It is thus computationally intensive at inference time (which may be problematic for real-time applications) and does not scale well with problem size.

We propose an intermediate method, referred to as a hybrid approach, which aims to capture the best of both worlds. The measured signal $\v{y}$ is projected onto a basis of $\Ntot{}$ fingerprints, pre-simulated using the biophysical model $S$. The naturally sparse representation (in theory, $K\ll \Ntot{}$ weights are non zero) is then fed to a neural network with an architecture driven by the physics of the process for final prediction of the latent biophysical parameters $\gv{\omega}$.

We describe the theory for general applications and provide an illustrative example of the estimation of brain tissue properties from diffusion-weighted magnetic resonance imaging (DW-MRI) data, where the inverse problem is learned on simulated data.
We compare the three methods (dictionary fingerprinting, hybrid and fully-learned) in terms of efficiency and accuracy on the estimated brain properties. The explainability of the hybrid and fully-learned approaches are assessed visually by projecting intermediate activations in a 2D space.

\section{Related work}
Despite their success at solving many (underdetermined) inverse problems~\cite{lucas2018using,bai2020deep} and despite efforts to explain model predictions in healthcare~\cite{elshawi2020interpretability,stiglic2020interpretability}, DNNs still offer little accountability and are mathematically prone to major instabilities~\cite{gottschling2020troublesome}. There has thus been a growing trend toward incorporating engineering and physical knowledge into deep learning frameworks~\cite{lucas2018using}. One way to do so is to unfold well-known, often iterative algorithms in a DNN architecture~\cite{ye2017tissue,ye2019deep}, or to produce parameter updates based on signal updates from the forward model $S$~\cite{ma2020deep}. However those approaches require many evaluations of the forward model $S$, which is not always possible or affordable. 

The latent variables $\gv{\omega}$ can be difficult to access. This is the case in our example problem where microstructural properties of the brain cannot be measured \textit{in vivo}. In such cases, DNNs can be trained on synthetic data $S\left(\gv{\omega};\mathcal{P}\right)$ and then applied to experimental measurements $\v{y}$, which is the approach taken with our illustrative example. Another option is via a self-supervised framework wherein the inverse mapping $S^{-1}$ is learned such that $S\left(S^{-1}\left(\v{y}\right)\right) \approx \v{y}$~\cite{senouf2019self}. However, the latter again requires many evaluations of $S$.

Our physics-based fingerprinting approach is based on the framework by~\cite{ma2013magnetic} for magnetic resonance fingerprinting (MRF) extended to the multi-contribution case~\cite{rensonnet2018assessing,rensonnet2019towards}. As dictionary sizes increased and inference via fingerprinting became slower, DNN methods have been proposed to accelerate the process~\cite{oksuz2019magnetic,golbabaee2019geometry}. However, these do not consider the case of multiple linear signal contributions as in Eq.~\eqref{eq:linearity_S}.

In the field of DW-MRI (our illustrative example), training of DNNs using high-quality data $\v{y}$, i.e. data acquired with a rich protocol $\mathcal{P}$, has been suggested in several works~\cite{golkov2016q,ye2017tissue,schwab2018joint,fang2019deep,ye2020improved}. The main limitation is that such enriched data may not always be available. It is worth noting that \cite{ye2020improved} also proposed Lasso bootstrap to quantify prediction uncertainty, as a way to strengthen interpretability.

\section{Methods}
\label{sec:methods}

\subsection{Running example: estimation of white matter microstructure from DW-MRI}
\label{sec:example}
In DW-MRI, $M$ volumes of the brain are acquired by applying magnetic gradients with different characteristics $\v{p}$ (intensity, duration, profile). In each voxel of the brain white matter, a measurement vector $\v{y}\in \mathbb{R}^M$ is obtained by compiling the $M$ DW-MRI values at this given voxel location. 

\begin{figure}
    \centering
    \includegraphics[width=0.5\textwidth, clip=True, trim=0cm 7.0cm 7.25cm 0cm]{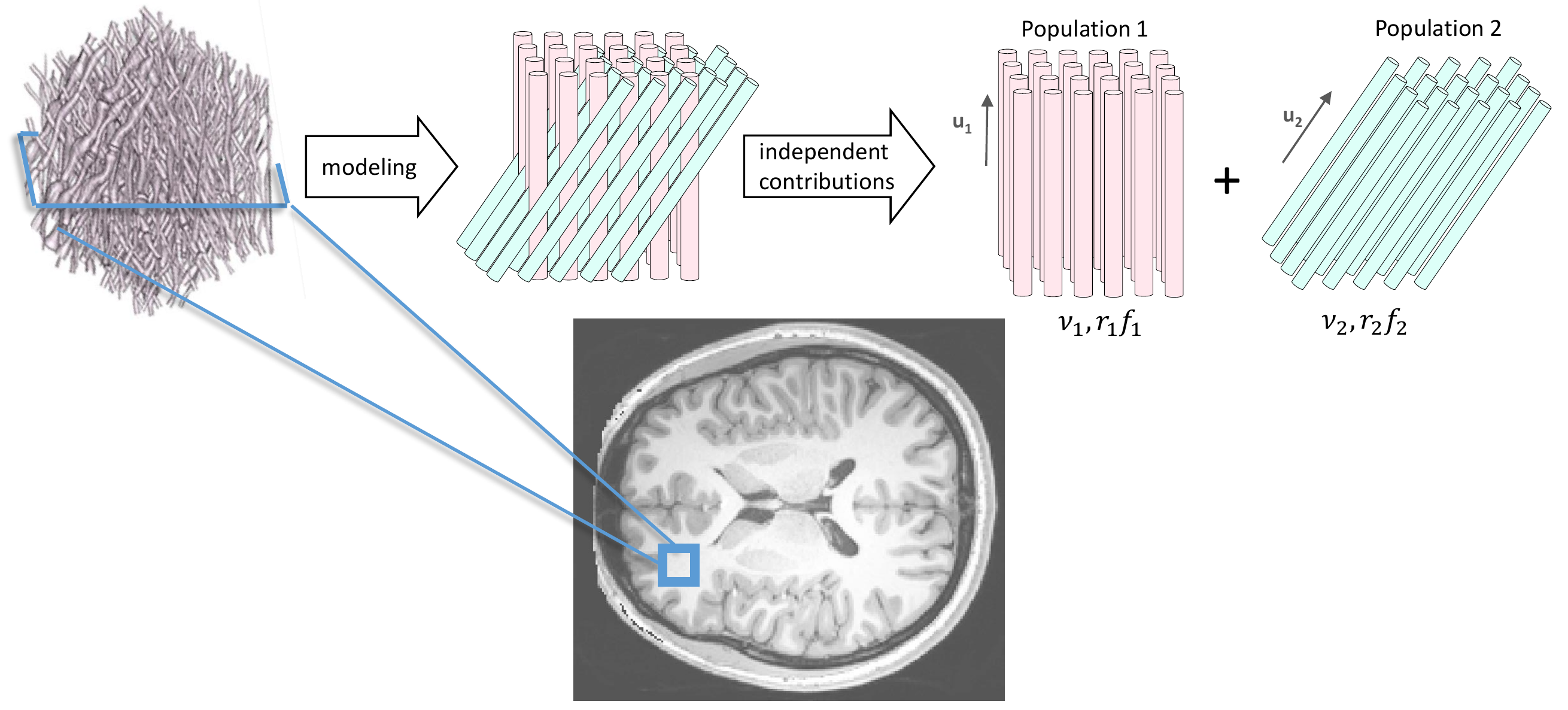}
    \caption{\textbf{Running example : estimation of white matter microstructure.} A voxel of white matter, shown here on a T1 anatomical scan of a healthy young adult from the Human Connectome Project \cite{van2012human}, contains crossing populations of roughly parallel axons. In our example model, each population is described by an orientation $\v{u}$, a volume fraction $\nu$, an axon radius index $r$ and an axon density index $f$ and contributes independently to the DW-MRI signal. The vector of biophysical parameters is $\gv{\omega}=\left[\v{u}_1^T,\nu_1, r_1, f_1, \v{u}_2^T, \nu_2, r_2, f_2\right]^T$. }
    \label{fig:microstructure}
\end{figure}

As illustrated in Figure~\ref{fig:microstructure}, the white matter is mainly composed of long, thin fibers known as \emph{axons} which tend to run parallel to other axons, forming bundles called \emph{fascicles} or \emph{populations}. In a voxel at current clinical resolution, 2 to 3 populations of axons may intersect \cite{jeurissen2013investigating}. In this paper a value of $K=2$ populations is assumed for simplicity. We use a simple model of a population of parallel axons (Figure~\ref{fig:microstructure}) parameterized by a main orientation $\v{u}$, an axon radius index $r$ (of the order of $\SI{1}{\micro\meter}$) and an axon fiber density $f$ (in $[0, 0.9]$), which are properties involved in a number of neurological and psychiatric disorders \cite{chalmers2005contributors, mito2018fibre, andica2020neurocognitive}. Assuming the signal contribution of each population $k$ to be independent~\cite{rensonnet2018assessing}, the generative signal model $S$ in a voxel is
\begin{equation}
S\left(\gv{\omega};\mathcal{P}\right) = \phi_{\textrm{SNR}}\left(\sum\limits_{k=1}^K \nu_k S_{\textrm{MC}}\underbrace{(\v{u}_k, r_k, f_k}_{\gv{\omega}_k};\mathcal{P})\right),
\label{eq:signal_model_dwmri}
\end{equation}
where the signal of each population is modeled by an accurate but computationally-intensive Monte Carlo simulation $S_{\textrm{MC}}$ \cite{hall2009convergence,rensonnet2015hybrid}. The weights $\nu_k$ can be interpreted as the fraction of voxel volume occupied by each population.  The model is stochastic with $\phi$ modeling corruption by Rician noise \cite{gudbjartsson1995rician} at a given signal-to-noise ratio (SNR). In most clinical settings, SNR is high enough for Rician noise to be well approximated by Gaussian noise. A least-squares data fidelity term $\left\|\v{y} - S\left(\gv{\omega};\mathcal{P}\right) \right\|_2^2 $ is thus popular in brain microstructure estimation as it corresponds to a maximum likelihood solution.

The acquisition protocol $\mathcal{P}$ is the MGH-USC Adult Diffusion protocol of the Human Connectome Project (HCP), which comprises $M=552$ measurements \cite{setsompop2013pushing}.

\subsection{Physics-based dictionary fingerprinting}
\label{sec:fingerprinting}
Fingerprinting is a general approach essentially consisting in a sparse look-up in a vast precomputed dictionary containing all possible physical scenarios. A least squares data fidelity term is assumed here~\cite{bai2020deep} although the framework could be extended to other objective functions.

For each contribution $k$ in Eq.~\eqref{eq:linearity_S}, a sub-dictionary $\v{C}^k \in\mathbb{R}^{M\times N_k}$ is presimulated once and for all by calling the known $S_k$, to form the total dictionary $\mathcal{D}\coloneqq \left[\v{C}^1 \dots \v{C}^K\right] \in \mathbb{R}^{M\times \Ntot{}}$, where $\Ntot{}\coloneqq \sum_{k=1}^K N_k$. Each $\v{C}^k$ thus contains $N_k$ atoms or \emph{fingerprints} $\v{A}^k_{j_k}\coloneqq S_k\left(\gv{\omega}_{kj_k};\mathcal{P}\right) \in \mathbb{R}^M $, with $j_k=1,\dots,N_k$, 
corresponding to a sampling of $N_k$ points in the space of biophysical parameters given a known experimental protocol $\mathcal{P}$. The dictionary look-up in this multi-contribution setting can be mathematically stated as
\begin{equation}
\begin{array}{lll}
\hat{\v{w}}= & \argmin\limits_{\v{w}\geq 0} 	
			& \left\|\v{y}-\begin{bmatrix}\v{C}^1 \dots \v{C}^K\end{bmatrix}\cdot
							\begin{bmatrix}\v{w}_1\\ \vdots \\ \v{w}_K\end{bmatrix}
																					\right\|_2^2\\ 
 & & \\
 & \text{subject to} & \left| \mathbf{w}_k \right|_{0}=1, \quad k=1,\dots,K, \\
\end{array}
\label{eq:sparse_optimization}
\end{equation}
where the sparsity constraints $\left|\cdot\right|_0 $ on the sub-vectors $\v{w}_k$ guarantee that only one fingerprint $\v{A}^k_{j_k}$ per sub-dictionary $\v{C}^k$ contributes to the reconstructed signal. No additional tunable regularization of the solution is needed as all the constraints (e.g., lower and upper bounds on latent parameters $\gv{\omega}$) are included in the dictionary $\mathcal{D}$. Equation~\eqref{eq:sparse_optimization} is solved exactly by exhaustive search, i.e. by selecting the optimal solution out of $\prod_{k=1}^K N_k$ independent non-negative linear least squares (NNLS) sub-problems of $K$ variables each
\begin{equation}
\begin{split}
(\hat{j}_1,\dots, \hat{j}_K)=& \\
\argmin\limits_{1\leq j_k \leq N_k}\quad & \min\limits_{\v{w}\geq 0} \left\| \v{y}-\begin{bmatrix}\v{A}^1_{j_1} \dots  \v{A}^K_{j_K}\end{bmatrix}\cdot
							\begin{bmatrix}w_1\\ \vdots \\ w_K \end{bmatrix}  \right\|_2^2.
\end{split}
\label{eq:combinatorial_optimization}
\end{equation}
Each sub-problem is convex and is solved exactly by an efficient in-house implementation\footnote{\texttt{solve\_exhaustive\_posweights} function of the Microstructure Fingerprinting library available at \url{https://github.com/rensonnetg/microstructure_fingerprinting}. Written in Python 3 and optimized with Numba (\url{http://numba.pydata.org/}).} of the active-set algorithm~\citep[][chap. 23, p. 161]{lawson1995solving}, which empirically runs in $\mathcal{O}\left(K\right)$ time. The optimal biophysical parameters $\hat{\v{\omega}}_k$ are finally obtained as those of the optimal fingerprint $\hat{j}_k$ in each $\v{C}^k$ and the weights $\hat{\nu}_k$ are estimated from the optimal $\hat{w}_k$.

The main limitation of this approach is its computational runtime complexity $\mathcal{O}\left(N_1\dots N_K K\right)$ or $\mathcal{O}\left(N^K K\right)$ if $N_k=N $ $\forall k$. Additionally, the size $N_k$ of each sub-dictionary $\v{C}^k$ also increases rapidly as the number of biophysical parameters, i.e. as the number of entries in $\gv{\omega}_k$ increases.

In our running example, $K=2$ sub-dictionaries of size $N=782$ each are pre-computed using Monte Carlo simulations, corresponding to biologically-informed values of axon radius $r$ and density $f$. The orientations $\v{u}_1, \v{u}_2$ are pre-estimated using an external routine \cite{tournier2007robust} and directly included in the dictionary.

\subsection{Fully-learned neural network}
\label{sec:full_nn}
\begin{figure*}[h!]
\centering
\includegraphics[width=0.60\textwidth, clip=true, trim=0cm 0cm 11.5cm 0cm]{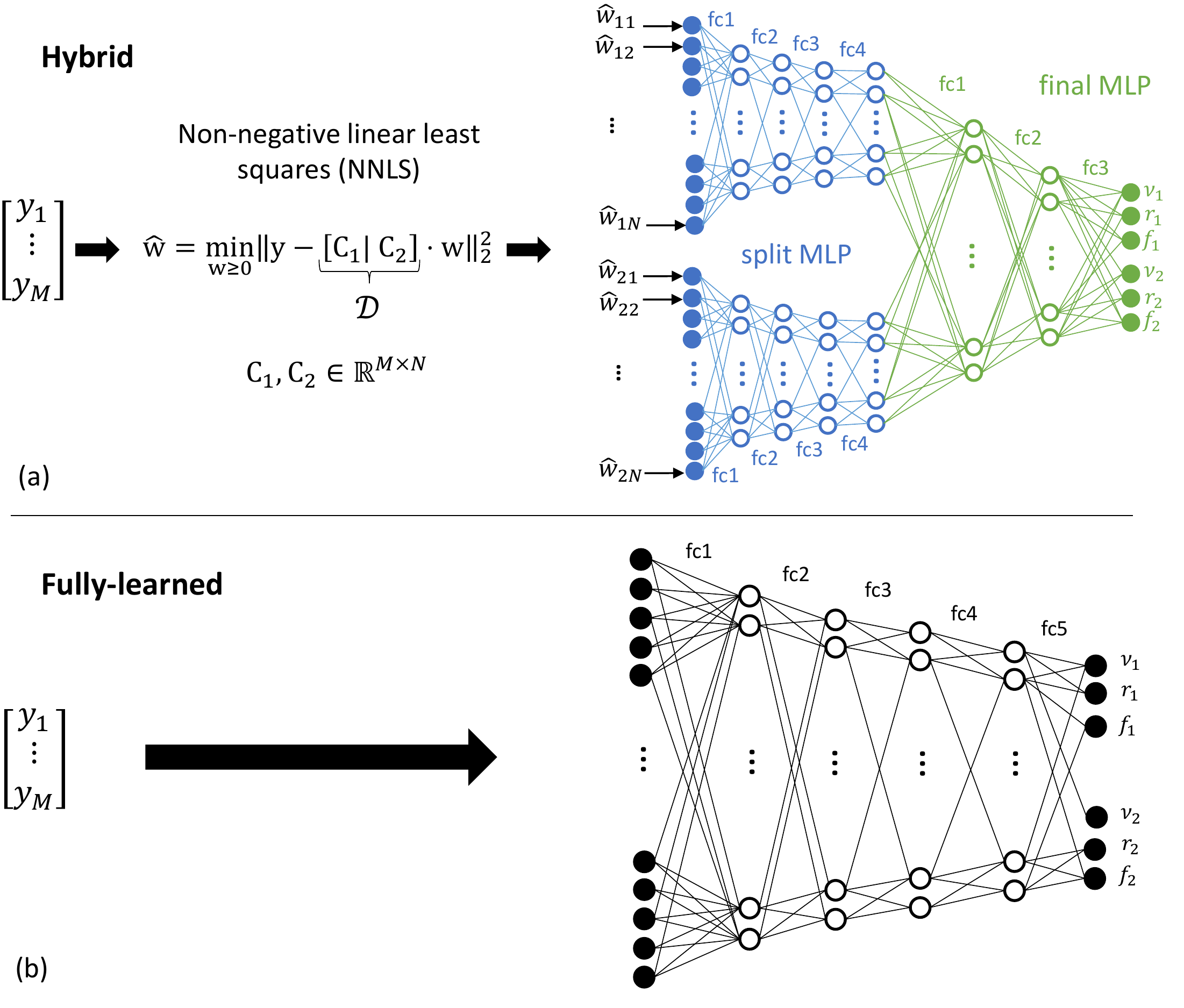}
\caption{\textbf{Interpretable hybrid vs fully-learned approach.} (a) The vector of measurements $\v{y}$ is decomposed by NNLS into a sparse representation in the space of physics-based fingerprints (i.e., many $w_{kj_k}$ are zero). The weights are given to a split multi-layer perceptron (MLP) followed by final fully-connected (fc) layers to predict the tissue parameters (axon radius $r$, axon fiber density $f$, relative volume of axon population $\nu$). (b) In the fully-learned approach, the input is directly fed to a DNN which makes the final prediction. Rectified linear units (ReLU) follow every fc layer in both networks.}
\label{fig:architectures}
\end{figure*}

As depicted in Figure~\ref{fig:architectures}, we focus on a multi-layer perceptron (MLP) architecture with rectified linear unit (ReLU) non-linearities (not shown in Figure~\ref{fig:architectures}), which has the benefit of a fast forward pass for inference. In our brain microstructure example, training is performed on simulated data obtained by Eq.~\eqref{eq:signal_model_dwmri} with tissue parameters $\gv{\omega}$ drawn uniformly from biophysically-realistic ranges. A mean squared error (MSE) loss is used to match Eq.~\eqref{eq:sparse_optimization} and \eqref{eq:combinatorial_optimization}. Dropout \cite{srivastava2014dropout} is included in every layer during training and stochastic gradient descent with adaptive gradient (Adagrad) \cite{duchi2011adaptive} is used to estimate parameters.

\subsection{Hybrid method}
\label{sec:hybrid}
The proposed method is a combination of the fingerprinting and the end-to-end DNN approaches presented above.

\paragraph{First stage: NNLS.} The same sub-dictionaries $\v{C}^k$ as in Section~\ref{sec:fingerprinting} are computed. Instead of solving Eq.~\eqref{eq:sparse_optimization} with 1-sparsity contraints on the sub-vectors $\v{w}_k$, a single NNLS problem is solved 
\begin{equation*}
\begin{array}{lll}
\hat{\v{w}}= & \argmin\limits_{\v{w}\geq 0} 	
			& \left\|\v{y}-\mathcal{D}\cdot \v{w} \right\|_2^2,\\ 
\end{array}
\end{equation*} 
where the optimization completely ignores the structure of the dictionary $\mathcal{D}=\begin{bmatrix}\v{C}^1 \dots  \v{C}^K\end{bmatrix} \in \mathbb{R}^{M\times \Ntot{}} $. Unlike in Eq.~\eqref{eq:combinatorial_optimization}, only one reasonably large NNLS problem with $\sum_{k=1}^KN_k$ variables is solved rather than many small problems with $K$ variables. The runtime complexity of this algorithm is $\mathcal{O}\left(\Ntot{}\right)$ in practice and typically yields sparse solutions~\citep{slawski2011sparse}. In fact, if the model $S$ used to generate the dictionary were perfect for the measurements $\v{y}$ we would have $\left|\v{w}\right|_0=K \ll \Ntot{}$. There is however no guarantee that the true latent fingerprints in $\mathcal{D}$ will be among those attributed non-zero weights by the NNLS optimization. However, we expect those selected fingerprints with non-zero weights to have underlying properties $\hat{\gv{\omega}}$ close to and informative of the true latent biophysical properties. The second stage of the method can be seen as finding the right combination of these pre-selected features for an accurate final prediction.

In our running example, as in the fingerprinting approach, the orientations $\v{u}_1, \v{u}_2$ are pre-estimated using an external routine \cite{tournier2007robust} and directly included in the dictionary.

\paragraph{Second stage: DNN.} 
The output $\hat{\v{w}}$ of the first-stage NNLS estimation is given to the neural network depicted in Figure~\ref{fig:architectures}. Its architecture exploits the multi-contribution nature of the problem: each sub-vector $\hat{\v{w}}_{k}$ of $\hat{\v{w}}$ is first processed by a ``split'' independent multi-layer perceptron (MLP) containing $N_k$ input units (blue in Figure~\ref{fig:architectures}). 
Splitting the input has the advantage of reducing the number of model parameters while accelerating the learning of compartment-specific features by preventing coadaptation of the model weights~\cite{hinton2012improving}. A joint MLP (green in Figure~\ref{fig:architectures}) performs the final prediction of biophysical parameters. The output of all fully-connected layers is passed through a ReLU activation. Being a feed-forward network, inference is very fast once trained. The overall computational complexity of the hybrid method is therefore dominated by the first NNLS stage.

\section{Experimental results}

\subsection{Efficiency}
Table~\ref{tab:performance} specifies the values obtained during the fine-tuning of the meta parameters of the DNNs used in the fully-learned and hybrid approaches. As predicted by theory, the hybrid method is an order of magnitude faster than the physics-based dictionary fingerprinting for inference. However its NNLS first stage makes it slower than the end-to-end DNN solution. Training times were similar but the fully-learned approach had the advantage of bypassing the precomputation of the dictionary.

\begin{table*}[t]
\caption{Meta parameters and efficiency of the three methods.}
\label{tab:performance}
\vskip 0.15in
\begin{center}
\begin{small}
\begin{sc}
\begin{tabular}{lccc}
\toprule
     & Fingerprinting & fully-learned & Hybrid \\
\midrule
minibatch size          & $\times$  &  $\num{5000}$  & $\num{5000}$  \\
dropout rate            & $\times$  &  $\num{0.05}$   & $\num{0.1}$ \\
learning rate           & $\times$  & $\num{5e-4}$    & $\num{1.5e-3}$\\
training samples        & $\times$  & $\num{4e5}$    & $\num{4e5}$ \\
hidden units            & $\times$  & $\num{3600}$    & $\num{1700}$       \\
parameters              &  $\times$ & $\num{3e6}$    & $\num{4.6e5}$ \\
precomputation time     & $\approx \SI{2}{\day}$ & $\times$ & $\approx \SI{2}{\day}$ \\
inference time/voxel     & $\SI{1.25}{\second}$ &   $\SI{1.02e-4}{\second}$ &    $\SI{1.45e-1}{\second} $   \\
inference complexity & $\BigO{N_1\dots N_K K}$ & $\BigO{1}$  & $\BigO{N_1+\dots +N_K}$ \\
\bottomrule
\end{tabular}
\end{sc}
\end{small}
\end{center}
\vskip -0.1in
\end{table*}

\subsection{Accuracy}
$\num{15000}$ samples $\left(\gv{\omega},S\left(\gv{\omega};\mathcal{P}\right)\right)$ (never seen by our DNNs during training) were simulated using Eq.\eqref{eq:signal_model_dwmri} of our DW-MRI example, with biophysical parameters in realistic ranges and SNR levels 25, 50 and 100. In order to test the robustness of the approaches to parameter uncertainty, two scenarios were tested for the fingerprinting and hybrid methods. First,  the population orientations $\v{u}_1,\v{u}_2$ were estimated using \citep{tournier2007robust} and therefore subject to errors. Second, the reference groundtruth orientations were directly included in the dictionary $\mathcal{D}$. This did not apply to the fully-learned model as it learned all parameters from the measurements $\v{y}$ directly.

\begin{figure}
    \centering
    \includegraphics[scale=0.38, clip=True, trim=0.60cm 3.0cm 0.6cm 3.1cm]{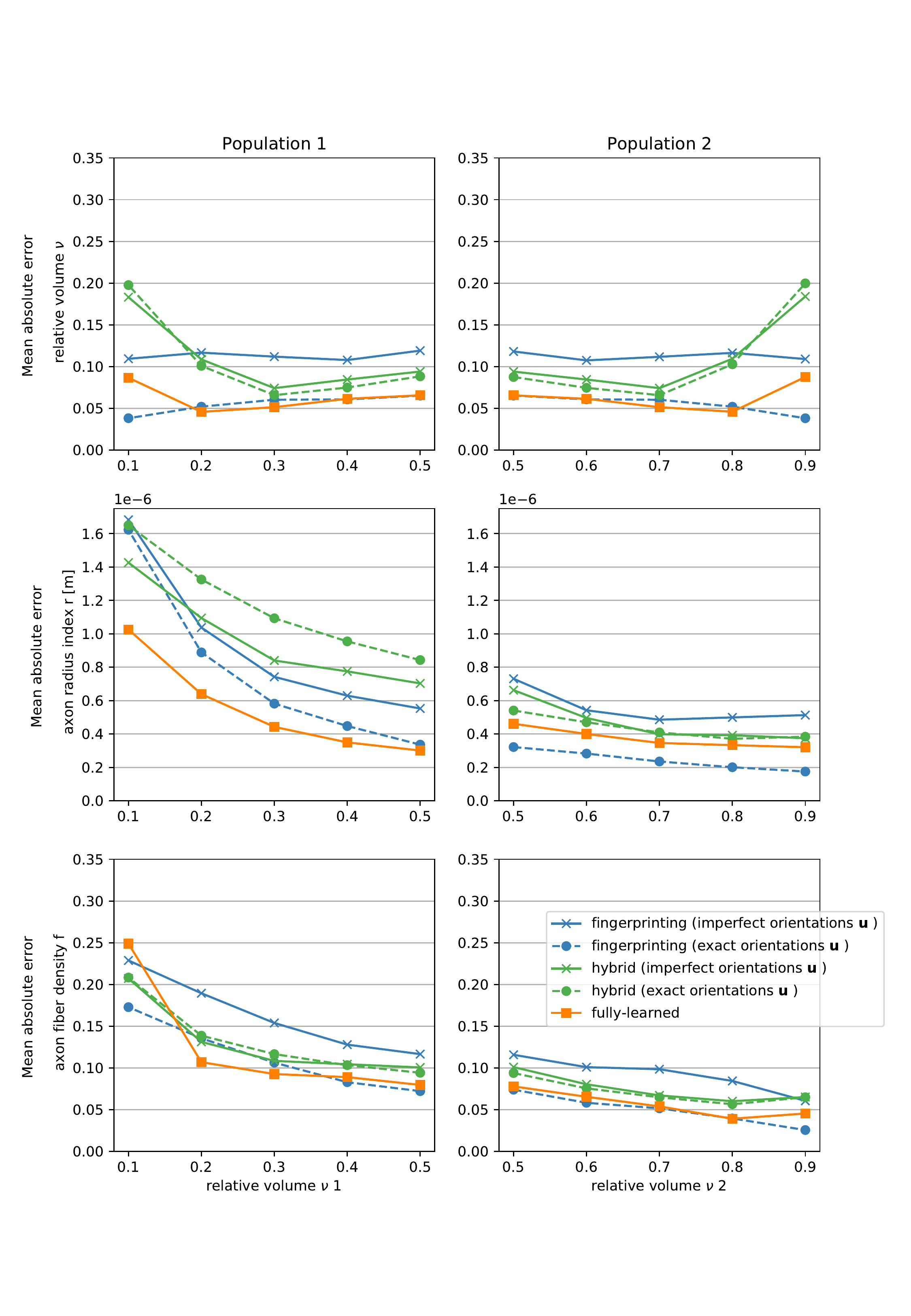}
    \caption{\textbf{The proposed hybrid method exhibits high accuracy and robustness.} Its accuracy on the estimated biophysical parameters measured by mean absolute error (MAE) from the reference groundtruth value was close to that of the end-to-end DNN approach for $\nu$ and $f$ and offered similar robustness to parameter uncertainty (here misestimated population orientations $\v{u}$). The estimation of $r$ (middle row) is notoriously difficult from DW-MRI data and should be interpreted with caution.}
    \label{fig:accuracy}
\end{figure}

Figure~\ref{fig:accuracy} shows that the hybrid approach (green lines) was more robust to uncertainty on $\v{u}$ than the physics-based fingerprinting method (blue lines), as the mean absolute errors (MAEs) only slightly increased when $\v{u}$ was misestimated (continuous lines). The fully-learned model exhibited the best overall performance. The poorer performance on the estimation of the radius index $r$ (middle row) is a well-known pitfall of DW-MRI as the signal only has limited sensitivity to $r$~\cite{clayden2015microstructural,sepehrband2016towards}. As the reference latent value of $\nu$ increased (x axis of Figure~\ref{fig:accuracy}), estimation of all tissue properties generally improved for all models.

\subsection{Explainability}
A total of $\num{11730}$ test samples were simulated using Eq.~\eqref{eq:signal_model_dwmri} in realistic ranges for $\nu,r,f$ and crossing angle between $\v{u}_1$ and $\v{u}_2$ at SNR 50. To ease visualization of the results, $r_1=r_2$ and $f_1=f_2$ was enforced. The activations, defined as the vector of output values of a fully-connected layer \emph{after} the ReLU activation, were inspected at different locations of the DNNs used in the hybrid and the fully-learned methods. The projection of these multi-dimensional vectors into a 2-dimensional space was performed using t-SNE embedding~\cite{van2008visualizing} for each of the $\num{11730}$ test samples. The idea of this low-dimensional projection was to conserve the inter-sample distances and topology of the high-dimensional space.

\begin{figure*}[h!]
\centering
\includegraphics[width=0.85\textwidth, clip=true, trim=0cm 2cm 0cm 0cm]{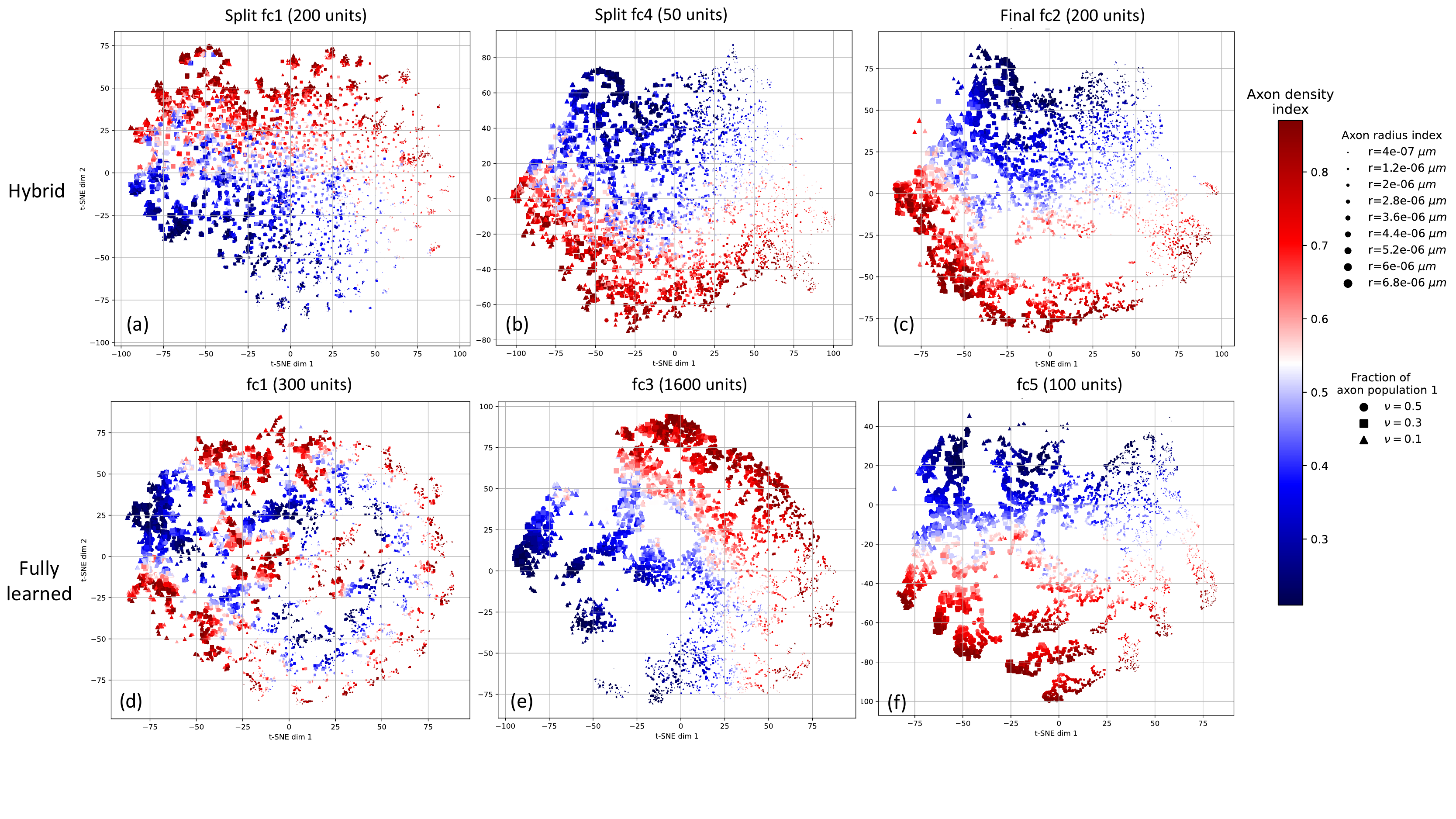}
\caption{\textbf{The neural network of the hybrid approach is faster to learn useful signal representations.} Projection in 2-dimensional plane of network activations using t-SNE embedding~\cite{van2008visualizing}. Top row: network used in the hybrid approach. Activations after (a) the first layer, (b) the last layer of the split MLP; (c) the last layer of the final MLP before final prediction. The input layer has already learned the axon radius $r$ and density $f$ and the final layers learn the relative volumes $\nu$ after the parallel networks merge (Figure~\ref{fig:architectures}). Bottom row: network in the fully-learned approach. Activations after (d) the first layer, (e) an intermediate layer, (f) the final layer. The visualization suggests that the first layer hasn't learned any tissue parameter yet.}
\label{fig:activations}
\end{figure*}

As shown in Figure~\ref{fig:activations}, the DNN in the second stage of the hybrid method seemed to learn the $r$ and $f$ properties after just the first layer, while it took the fully-learned model three layers to display a similar sample topology. At the end of the split MLP the $\nu$ parameter (marker shapes in Figure~\ref{fig:activations}) did not seem to have been learned but the different values of $\nu$ were then well separated at the end of the merged final MLP (see architecture in Figure~\ref{fig:architectures}).

\section{Discussion and conclusion}
An approach was proposed combining the efficiency of deep neural networks with the interpretability of traditional optimization for general inverse problems in which evaluation of the forward generative model $S$ is possible but expensive. The potential of the method was exemplified on a problem of white matter microstructure estimation from DW-MRI. The DNN used in the hybrid method receives as input the measurements $\v{y}$ expressed as a linear combination of a few representative fingerprints taken from a physically-realistic basis. Exploiting the multi-contribution nature of the physical process, this input is separated by signal contribution and fed to independent MLP branches (see Figure~\ref{fig:architectures}). The overall effect is an expedited learning process, as suggested in Figure~\ref{fig:activations} where latent parameters $\v{\omega}_k$ are learned directly, ``for free'', for each axon population $k$. Consequently, we could consider further reducing the number and/or size of layers in the split MLP part of the network used in our illustrative example.

Figure~\ref{fig:activations} also confirms that the weights $\nu$ of the independent signal contributions are only learned in the final, joint MLP of the DNN of the hybrid method. This is because a relative signal weight can only be defined with respect to the \emph{other} weights. The split branches of the network treat individual signal contributions independently, unaware of the other contributions, and thus cannot learn the relative weights $\nu$. This suggests that they more specifically focus on the properties $\v{\omega}_k$ of each contribution.

A limitation of the proposed hybrid approach is the need to simulate the dictionary $\mathcal{D}$ which can be costly as the number of latent biophysical parameters increases. With high-dimensional inputs such as medical images, it may also be necessary to consider better adapted convolutional network architectures in the second stage of the hybrid method. General decompositions such as PCA analysis could then be considered~\cite{harkonen2020ganspace}. Our exemplary experiments were performed on simulated data and further investigation is required to demonstrate that the method generalizes to experimental data.

Further work should also test whether the hybrid approach enables transfer learning from a protocol $\mathcal{P}$ to a new protocol $\mathcal{P}'$. In our DW-MRI example, this could happen with a scanner update or protocol shortened to accomodate pediatric imaging, for instance. While a new dictionary $\mathcal{D}'$ would be required for the fingerprinting and the hybrid method, the DNN trained in the hybrid approach should in theory still perform well without retraining. This is because fingerprints in $\mathcal{D}'$ would be linked to the same latent biophysical parameters $\gv{\omega}_{kj_k}$ (the number of rows $M'$ of $\mathcal{D}'$ would change, but not the number of columns $\Ntot{}$). Preliminary results (not shown in this paper) were encouraging. The DNN of the fully-learned approach would need to be retrained completely however. 

Future work will also more closely inspect the inference process in the network via techniques such as LIME~\cite{ribeiro2016should}, guided back-propagation~\cite{selvaraju2017grad}, shapley values and derivatives~\cite{shapley201617,sundararajan2020many} or layer-wise relevance propagation~\cite{bohle2019layer}. This would complement our t-SNE inspection and further reinforce the confidence in the prediction of our approach.

We hope that these preliminary findings may contribute to incorporating more domain knowledge in deep learning models and ultimately encourage the more widespread adoption of machine learning solutions in the medical field.

\medskip

\FloatBarrier


\section*{Acknowledgements}
Computational resources have been provided by the supercomputing facilities of the Universit\'{e} catholique de Louvain (CISM/UCL) and the Consortium des \'{E}quipements de Calcul Intensif en F\'{e}d\'{e}ration Wallonie Bruxelles (C\'{E}CI) funded by the Fond de la Recherche Scientifique de Belgique (F.R.S.-FNRS) under convention 2.5020.11 and by the Walloon Region.
%


\bibliography{bibliography}
\bibliographystyle{icml2021}

\end{document}